\documentclass[pdflatex,sn-mathphys]{sn-jnl}

\renewenvironment{table}[1][]%
{\tableorg[#1]%
\tablebodyfont%
\renewcommand\footnotetext[2][]{{\removelastskip\vskip3pt%
\let\tablebodyfont\tablefootnotefont%
\hskip0pt\if!##1!\else{\smash{$^{##1}$}}\fi##2\par}}%
}{\endtableorg}

\usepackage{bookmark}
\usepackage{algorithm}
\usepackage{algpseudocode}
\usepackage{parskip}
\usepackage{float}
\usepackage{pdfpages}
\usepackage{multirow}
\usepackage{lscape}
\usepackage{longtable}
\usepackage{booktabs}
\usepackage{listings}
\usepackage{xcolor}
\usepackage[utf8]{inputenc}

\definecolor{codegreen}{rgb}{0,0.6,0}
\definecolor{codegray}{rgb}{0.5,0.5,0.5}
\definecolor{codepurple}{rgb}{0.58,0,0.82}
\definecolor{backcolour}{rgb}{0.95,0.95,0.92}

\lstdefinestyle{mystyle}{
    backgroundcolor=\color{backcolour},   
    commentstyle=\color{codegreen},
    keywordstyle=\color{magenta},
    numberstyle=\tiny\color{codegray},
    stringstyle=\color{codepurple},
    basicstyle=\ttfamily\footnotesize,
    breakatwhitespace=false,         
    breaklines=true,                 
    captionpos=b,                    
    keepspaces=true,                 
    numbers=left,                    
    numbersep=5pt,                  
    showspaces=false,                
    showstringspaces=false,
    showtabs=false,                  
    tabsize=2
}

\jyear{2023}%

\theoremstyle{thmstyleone}%
\newtheorem{theorem}{Theorem}
\newtheorem{proposition}[theorem]{Proposition}%

\theoremstyle{thmstyletwo}%

\theoremstyle{thmstylethree}%
\newtheorem{definition}{Definition}%

\MakeRobust{\Call}
\begin{document}

\title[Explanation Comparison and Implication on Model Performance]{The XAISuite framework and the implications of explanatory system dissonance}


\author*[1]{\fnm{Shreyan} \sur{Mitra}}\email{shreyan.m.mitra@gmail.com}

\author[2]{\fnm{Leilani} \sur{Gilpin}}

\affil*[1]{\orgname{Adrian C. Wilcox High School}, \orgaddress{\street{3250 Monroe Street}, \city{Santa Clara}, \postcode{95051}, \state{California}, \country{United States}}}

\affil[2]{\orgdiv{Department of Computer Science}, \orgname{UC Santa Cruz}, \orgaddress{\street{1156 High St}, \city{Santa Cruz}, \postcode{95064}, \state{California}, \country{United States}}}


\abstract{
	Explanatory systems make machine learning models more transparent. However, they are often inconsistent. In order to quantify and isolate possible scenarios leading to this discrepancy, this paper compares two explanatory systems, SHAP and LIME, based on the correlation of their respective importance scores using 14 machine learning models (7 regression and 7 classification) and 4 tabular datasets (2 regression and 2 classification). We make two novel findings. Firstly, the magnitude of importance is not significant in explanation consistency. The correlations between SHAP and LIME importance scores for the most important features may or may not be more variable than the correlation between SHAP and LIME importance scores averaged across all features. 
Secondly, the similarity between SHAP and LIME importance scores cannot predict model accuracy. In the process of our research, we construct an open-source library, XAISuite, that unifies the process of training and explaining models. Finally, this paper contributes a generalized framework to better explain machine learning models and optimize their performance.}




\maketitle

\section{Introduction}\label{sec1}

From self-driving cars to customer support chat-bots, machine learning models have become pervasive in our daily lives. \cite{nirmal_2017} The problem is that these machine learning models are opaque - that is, the mechanisms by which a model arrives at a particular result are not known by humans.  When these opaque systems are being entrusted with human-level decision, e.g., handing down sentences to convicts or driving a car, they will need to be able to explain themselves to justify their behavior. \cite{lakshmanan_2021}

This is especially pertinent when such opaque models fail. In 2016, a ProPublica article revealed that Northpointe, a widely used criminal risk assessment tool, incorrectly rated incarcerated African Americans as more likely to commit future crimes than Caucasians. \cite{angwin_larson_kirchner_mattu_2016} And in 2018, a self-driving car hit and tragically killed a cyclist. \cite{wakabayashi2018self} The machine learning model in the car was unable to reconcile contrasting information from various sensors, and thus failed to make the right decision. \cite{jones_2018}

Explanatory systems produce explanations, or model-dependent justifications. \cite{rose2019} They provide one way to understand machine learning models. However, for explanations to be trustworthy, it is essential that they are consistent and accurate. \cite{gilpin_2018} We define consistency as having two components. The first part is reproducibility - applying explanation methods repeatedly should yield the same results. Secondly, the results of different explanatory methods using the same dataset and model should be similar. By accurate, we mean that the justifications provided by explanation systems are correct. In this paper, we examine two state of the art explanatory systems: SHAP \cite{NIPS2017_7062}, based on the game theory concept of Shapley values, and LIME \cite{lime}, which stands for local interpretable model-agnostic explanations. We propose an approach to compare SHAP and LIME explanations and to automatically analyze cases where they are inconsistent. 

Furthermore, by highlighting the inconsistencies between explanatory systems, this paper contributes to making explanations for machine learning models more consistent and accurate. This will provide users and stakeholders a supported reason behind system malfunctions, preventing incidents like the one involving the self-driving car and the cyclist. In addition, system debugging and diagnosis will be more efficient. 

Explanations that users can trust are essential. Without trustworthy explanations, we are effectively blind to the operation of machine learning models and cannot mitigate their flaws. Our work answers the following research questions: 

\begin{itemize}
	\item How can users trust explanatory systems? 
	\item How can accuracy of explanation methods be measured? 
	\item And, finally, how can we develop such trustworthy explainers for machine learning? 
\end{itemize}

In this paper, we present a XAISuite framework that attempts to answer these questions. 

XAISuite provides an interface to compare different explanatory systems. Users can trust explanations more if they see that there is a consensus among different explanatory systems. 

Using the XAISuite framework, our research concludes that correlation between explanatory systems is not necessarily associated with the performance of a machine learning model. By showing that explanatory systems may agree even in cases where the model fails, we caution users of the accuracy of explanation methods in such cases and open research into other possible indicators of explanation accuracy. 

And as we seek to progress towards more trustworthy explainers, we expect the results of different explanatory systems to converge to a single correct justification of a particular model's performance. XAISuite's explanation generation and comparison utilities will help in this process by highlighting cases where different explanatory systems differ. 

\section{Background/Related Work}\label {sec2}

Our paper builds on previous work in explanations, failure analysis, and machine learning error.

\textbf{Bases for Explanation Accuracy} 

In the introduction, we defined explainer accuracy as how similar the justifications produced by explanation systems are to reality. But what is the baseline by which reality is defined?

A research paper published by DeepMind \cite{https://doi.org/10.48550/arxiv.1706.08606} suggests that the answer lies in human thought. The paper explores the use of cognitive psychology to explain the decisions of machine learning models, drawing parallels between biases humans develop during their maturation and those acquired by machine learning. By likening machine learning models to humans, the paper provides a framework to determine which explanations have a higher probability of being accurate. Since explanations are ultimately meant for human understanding, we find the use of psychology in explanation generation promising and perhaps capable of resolving the discrepancies between explanatory systems outlined in this paper. 

Gilpin et al. \cite{gilpin_2018} believe that what is defined as accurate might depend on user requirements. They note a tradeoff between completeness and interpretability that all explanatory systems must follow - the more accurate explanations are, the less likely they are to be understandable by humans. This tradeoff may affect the discrepancies between different explanatory models. Thus, a key part of future explanatory system research is creating explainers that gain the user's trust. \cite{goel_sindhgatta_kalra_goel_mutreja_2022} We see our work in proposing an automated system to ensure the consistency and accuracy of explanatory systems as essential to that effort. Research on user requirements for explanations is elaborated on further later in this section.

Han et al. \cite{han2022} propose that different explanatory systems are optimal for different scenarios, and an ``adversarial" sample exists that will lead to a large error for any given explanatory system. For example, while SHAP and LIME are both based on local function approximation (LFA), they differ in their optimal intervals due to their noise functions. This can help explain the discrepancies between SHAP and LIME that we observe in our research. It also suggests that the ``perfect" explainer can be created by joining different explainers over many different corresponding optimal intervals. Thus, if the correlations between SHAP and LIME found in this paper indicate that SHAP performs poorly where LIME performs well, and vice versa, it would be a big step towards improving explainer accuracy. Research on explanation comparison is explored later in this section.

\textbf{On user requirements for explanations}

Numerous papers \cite{riveiro2021} \cite{10.1145/1943403.1943424} \cite{10.1145/3503252.3531306} have highlighted the importance of user expectations in explanation utility. Since user expectations are often implicit, determining what type of explanations users are looking for is difficult. The XAISuite framework that we propose in this paper attempts to alleviate this problem by providing users with the option to use multiple explanatory systems, compare them, and choose the explanations most suitable to their scenario. 

Chazette et al. \cite{9920064} go further to propose and test a framework for explanatory systems focused on usage frequency and user frequency. We design XAISuite keeping the requirements outlined in the paper in mind.  

If machine learning is to be used in lieu of human-level decision making, machine learning models need to be safe and trustworthy \cite{10.1007/978-3-642-32378-2_8}. One way to ensure safety is to have stricter requirements and guarantees. In 2021, Nadia Burkart and Marco Huber \cite{burkart_2021} laid out the requirements of explainable supervised machine learning models. Our paper implements two of their requirements: (1) We make explanation of machine learning models easier and more accessible through our open source XAISuite library and (2) We ensure explanatory methods are consistent and easy to follow for humans.

Sometimes user requirements for explanations may vary in specialized fields. Ghassemi et al. \cite{Ghassemi2021} argue against the use of explainers in the medical profession, claiming that the many failures of explanatory systems endanger the trust of healthcare professionals and the lives of patients. They propose that machine learning models be rigorously tested instead. However, we believe that the solution to explainer error is not abandoning explanatory systems altogether, but to improve them until they are trustworthy. XAISuite is an attempt in that direction. 




\noindent \textbf{On comparing explanatory systems}

Comparing explanatory models is an open area of research. This is sometimes known as the ``disagreement problem" \cite{https://doi.org/10.48550/arxiv.2202.01602}. 

In their paper, Covert et al. \cite{covert2022} point out that while there are many different explanatory methods, it remains unknown how ``most methods are related or when one method is preferable to another." The authors propose a new class of similar explanations supported by cognitive psychology called removal-based explanations. These systems determine the importance of a feature by analyzing the impact of its removal. The paper specifically highlights that as SHAP and LIME are both part of the removal-based explanatory framework, they share a resemblance. The discrepancies between SHAP and LIME shown in our paper therefore have added value as markers of where two very similar explanatory systems with related internal mechanisms can differ. 

Roy et al. \cite{9978217} in ``Why Don't XAI Techniques Agree?" acknowledged that SHAP and LIME explanations often disagree and that users don't know which one to trust. They proposed an aggregate explainer that focuses on the similarities between SHAP and LIME and disregarded discrepancies. But the authors of that paper do not set forth a way to find and resolve the discrepancies. We contribute a
method to empower users to better understand the reasons to trust
SHAP and LIME.

van der Waa et al. \cite{VANDERWAA2021103404} extended explanatory system comparison further with a detailed analysis of rule-based versus example-based explanations, with implications on user trust and accuracy. Our framework allows users to validate this analysis by seeing for themselves the difference between different types of explanatory systems rather than having to trust only one of them. 

Duell et al. \cite{9508618} specifically compares the results of explanatory systems such as SHAP and LIME on electronic health records. They note significant differences in importance scores between the explanatory systems, stating that ``studied XAI methods circumstantially generate different top features; their aberrations in shared feature importance merit further exploration from domain-experts to evaluate human trust towards XAI." While this aligns perfectly with the results of our paper, we extend the study to different types of data outside of health records. We also create a generalized framework to help in the "further exploration" Duell et al. deemed necessary. 

The in-depth comparison that we perform between two explanatory systems has been previously explored. A paper published by Lee et al. \cite{Lee2022} compares breakDown (BD) and SHAP explainers in the specific case of classification of multi-principal element alloys. However, their work is not generalizable to all tabular data and all machine learning models, which XAISuite is. That paper also does not delve deeply into specific circumstances causing the differences between SHAP and BD explanations, which we do. In our work, we use various models on data of different types to allow us to have a better picture of exactly what factors affect explainer consistency. Furthermore, we compare SHAP and LIME, two of the most commonly used machine learning models in the field, and thus our results are applicable to more applications of machine learning. 

\noindent \textbf{Related Software}

Various tools similar to XAISuite also exist:

\begin{enumerate}
	\item{Agarwal et al. \cite{agarwal2022openxai} created a tool, OpenXAI, for evaluating and benchmarking post-hoc explanation systems, comparable in functionality and user interface to our XAISuite. While OpenXAI focuses more on accuracy of explanatory systems over one another in specific tasks, we put a heavier emphasis on consistency for all tasks. Furthermore, we set forth a fully-interactive graphical user interface that requires no code from the user-end.}
	
	\item{Yang et al. \cite{wenzhuo2022-omnixai} created the OmniXAI library for explainable AI that allows easy access to numerous explainers for a particular machine learning model. The OmniXAI library serves as part of the backend of the XAISuite library by helping to fetch explanatory systems that the user requests.}
	
	\item{Captum, similar to OmniXAI, was proposed by Kohklikyan et al. \cite{2009.07896}. In their implementation of the XAISuite library, the authors believed that OmniXAI was more compatible and easier to work with, but any future library that follows the XAISuite framework presented in this paper could use Captum to fetch explanatory importance scores.}
\end{enumerate}

\section{Methods}\label{sec11}

We divide the methodology into two parts: (1) creating the XAISuite framework and (2) using the framework to compare SHAP and LIME values, an example of which is shown in Fig.1. The latter part consists of two observational studies that constitute the basis of our findings. One concludes that correlation between SHAP and LIME importance scores cannot be used to predict model accuracy. The other shows that the correlations between SHAP and LIME importance scores for the most important features are consistently more variable than the correlation between average SHAP and LIME importance scores across all features, and therefore are not as reliable of a metric to test explainer similarity.

\begin{figure}[H]
	\centering
	\includegraphics[width=0.6\textwidth]{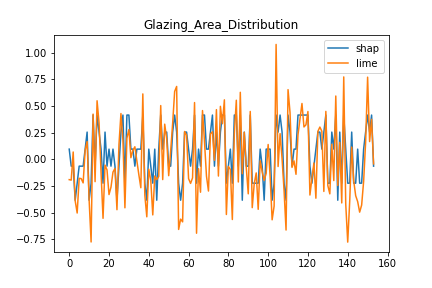}
	\caption{A comparative line chart of SHAP and LIME importance scores for Glazing Area as a feature in the UC Irvine Energy Efficiency Dataset using a Bayesian Ridge model. Graphs like this depict the micro-scale for explanations. Put together across many datasets and models, they can generate key insights about explainer accuracy and machine learning model performance.}
	\label{fig:examplecorrplot1}
\end{figure}







\subsection{Framework for training and explaining models}

We now present the XAISuite framework, a unified tool for comparing explanatory systems. It forms the basis of the XAISuite library, which enables users to train and explain models with minimal input. In constructing this library, we build on OmniXAI \cite{wenzhuo2022-omnixai}, which allows direct access to different explainers and helps us save the importance values generated by these explainers. 

A brief overview of the XAISuite framework is presented on the next page. The framework consists of three components: data retrieval, machine learning model creation and training, and explanation generation. A more detailed version with implementation suggestions is found in Appendix~\ref{secA1}. 

\begin{figure}[H]
	\caption{A brief flowchart of the XAISuite framework, drawn by the authors. The framework supports the XAISuite library, available at \url{https://github.com/11301858/XAISuite} The internal data manipulation and explanation generation in that library are courtesy of \cite{wenzhuo2022-omnixai} from Salesforce. From now on, when we refer to just ``XAISuite", we are referring to the framework, not the library.}
\end{figure}

\includepdf[scale=0.8, pages=1]{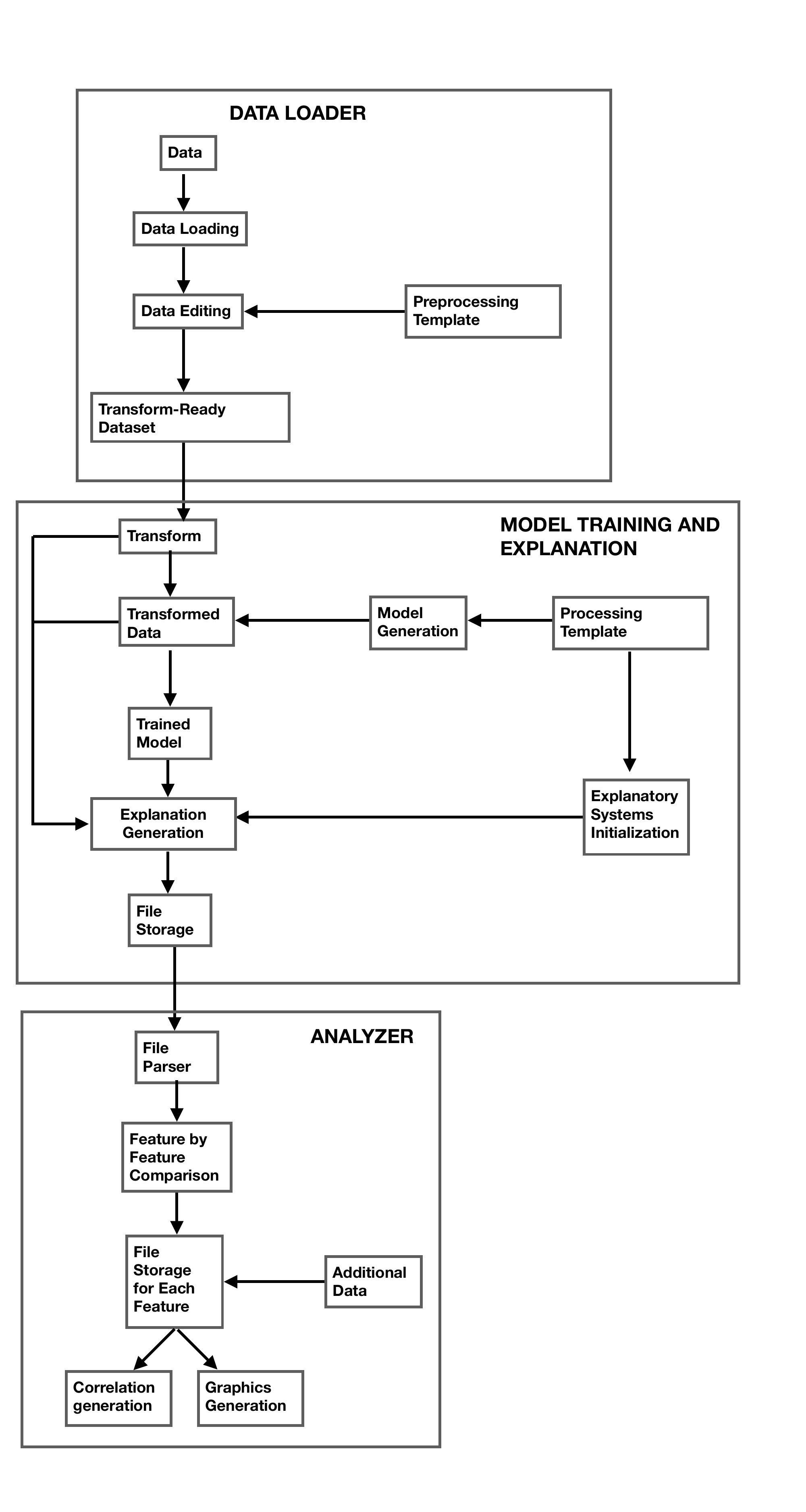}

We contribute the XAISuite framework with the intention for it to be a standard platform for training, explaining, and analyzing models. The framework was constructed with an eye on five key factors:

\begin{enumerate}
	\item \textbf{Simplicity: } Containing just three parts which depend on data retrieval, function calls, and writing to output files, XAISuite provides guiding principles for any implementation to enhance code changes and user convenience.
	\item \textbf{Integratablity: } A library is limited if it cannot be used with other libraries. Core functionalities, like model training, explanation generation, and graphics creation, are designed to use external libraries. However, the framework is flexible and not based on any specific external dependency, so there is flexibility in the way in which the model is trained or explanations are generated.
	\item \textbf{Flexibility: } A key feature of XAISuite is flexibility. This is enabled by the lack of specifics and the use of general terms. Note that \emph{any} dataset can be used, \emph{any} model can be trained, \emph{any} explanatory system can be initialized, depending on the implementation. The processing templates have no fixed form, nor is the form of data storage specified. As mentioned in the previous point, the XAISuite framework is compatible with any potential provider of its constituent parts, whether they be model libraries(sk-learn, XGBoost, etc.), transform function types (Logarithmic, Exponential, etc.), or different data storage options (Dataframe, Numpy, Files, etc.).
	\item \textbf{Usability: } Users are the center of explainability research \cite{riveiro2021}, and so any interface that facilitates interactions between the user and explanatory system must be user-centric. XAISuite achieves this by ensuring that results are understandable in a readable table or graphical format. Again, individual implementations of data or graph generation may vary, but by enforcing the requirement of converting data into a portable and visualizable medium, XAISuite reinforces the human-centric approach that is a hallmark of explainability research.
	\item \textbf{Expandability: } XAISuite is not designed to be a closed system. There are ample spaces provided for users to extend existing functionalities or link XAISuite with other existing frameworks. For example, the additional data that can be inserted at the Analyzer stage can be generated by other frameworks or modules. In fact, the XAISuite library provides two points of data generation that can be used at this stage.
\end{enumerate}

Algorithms and code for the implementation of the XAISuite Framework's machine learning model training and explanation utilities is included in Appendix ~\ref{secA2}. 

\subsection{Using the XAISuite framework to compare SHAP and LIME explanations}
Here, we describe the process of using the XAISuite framework and library to achieve the results detailed in this paper. The results themselves are in the Results section and the interpretation of the results are in the Discussion/Conclusions section. 

First, let's start with some definitions that will be further elaborated on in later sections:

\definition{\textbf{: First-Level Model} A model that is trained and explained using a given dataset.}
\definition{\textbf{: Second-Level Model} A model that is trained and explained on the explanatory importance scores, accuracy, or other output of a First-Level model. A Second-Level Model does not need to be the same type of model as the corresponding First-Level Model.}

\subsubsection{Preliminary Steps}
The following steps should be followed before moving onto any of the specific procedures in the following sections:

\begin{sidewaystable}
	\sidewaystablefn%
	\begin{center}
		\begin{minipage}{\textheight}
			\caption{Datasets used and corresponding machine learning models}\label{tab2}
			\begin{tabular*}{\textwidth}{@{\extracolsep{\fill}}lcccccc@{\extracolsep{\fill}}}
				\toprule%
				\textbf{Data}\footnote{For each task, we use a relatively simple dataset and complex dataset to account for the effect of data complexity in model training.} & \textbf{Type} & \textbf{Models}\footnote{Selected for variety in complexity from list of available supervised sk-learn models: \hyperlink{https://scikit-learn.org/stable/supervised_learning.html}{See source here}} & \textbf{Source} \\ \hline
				\midrule
				Diabetes          & \multirow{2}{*}{Regression}     & \multirow{2}{*}{\begin{tabular}[c]{@{}l@{}}Linear Regression, SGD Regressor, Kernel Ridge\\ Elastic Net, Bayesian Ridge, Gradient Boosting Regressor, \\ SVR\end{tabular}}                                              & Sk-learn                \\ \\ 
				Energy Efficiency &                                 &                                                                                                                                                                                                                         & UC Irvine ML Repository \\ \\ \cline{1-4}%
				Iris              & \multirow{2}{*}{Classification} & \multirow{2}{*}{\begin{tabular}[c]{@{}l@{}}Logistic Regression, Gaussian NB, K Neighbors Classifier, \\ Decision Tree Classifier, Random Forest Classifier\\ Gradient Boosting Classifier, Multinomial NB\end{tabular}} & Sk-learn                \\ \\
				Digits            &                                 &                                                                                                                                                                                                                         & Sk-learn               \\ \\ \cline{1-4} 
				\botrule
			\end{tabular*}
			\footnotetext{4 datasets were used, with two used for regression and the other two for classification tasks. Each type of task used 7 models for a total of 14 models used in the study.}
			
		\end{minipage}
	\end{center}
\end{sidewaystable}

\emph{For the steps below and in the steps in the following sections, we use n = 10, p = 0.5, and m = 5.}

\begin{enumerate}
	\item Install and import the XAISuite library \footnote{Documentation and Installation Directions found here: \hyperlink{https://11301858.github.io/XAISuite/}{XAISuite Page}}
	\item Import the necessary datasets. Refer to Table 1 (next page). For CSV files, download the file and place it in the current working directory. For sk-learn datasets, simply import the sklearn.datasets module
	\item Use the XAISuite library to train and explain the models listed in Table 1 for each dataset using SHAP and LIME. \footnote{For the Energy Efficiency Dataset, the authors chose \emph{Heating Load} as the target variable and disregarded \emph{Cooling Load}}. These models are our first-level models. Repeat this step \emph{n} times and calculate (1) the average SHAP-LIME correlation across all features of a dataset per model and (2) accuracy of these models on each dataset.
\end{enumerate}

The preliminary steps can be executed by the setup script in Appendix ~\ref{secA4}.

\subsubsection{Implications of correlation between SHAP and LIME on model accuracy}
We hypothesize that for higher disagreement between SHAP and LIME explanatory systems , i.e. lower correlations between SHAP and LIME importance scores, there will be more room for error and the machine learning model will be less accurate. Conversely, when there is higher agreement between SHAP and LIME explanatory systems, the model will be more accurate. We test our hypothesis by following the procedure below after execution of the preliminary steps:

\begin{enumerate}
	\item For each dataset, use  the 7 regression models listed in Table 1 (Linear Regression, SGD Regressor, Kernel Ridge, Elastic Net, Bayesian Ridge, Gradient Boosting Regressor, and SVR) with XAISuite to predict first-level model accuracy from SHAP and LIME correlation, as computed in the preliminary steps. Record the performance scores of these second-level models. Values above \emph{p} are taken to support the hypothesis. \footnote{We understand that low accuracies of second-level models could be attributed to the small size of the dataset generated from the first-level models (there are only 7 instances because there are 7 models). We leave it to others to determine if our results hold for larger training sizes for second-level models, i.e. use of more first-level models, and for non-tabular datasets.}
\end{enumerate}
Repeat the preliminary steps and the above step \emph{m} times to ensure consistency of results. Recall from the preliminary steps that we took \emph{m} = 5.

We now present a possible algorithm to implement this step in Pythonic language, but this algorithm can be generalized to support other programming languages as well. Implementations of the \Call{train}{}, \Call{Corr}{}, and \Call{getExplanations}{} functions are not shown because they are not contributions of this paper. \footnote{In the case of Python, \Call{Train}{} is implemented in sklearn, \Call{Corr}{} is implemented in pandas, and \Call{getExplanations}{} is implemented in OmniXAI.}

\begin{algorithm}[!htbp]
	\scriptsize{}
	\caption{Use SHAP and LIME correlations to predict model accuracy}\label{algo1}
	\begin{algorithmic}
		\Function{trainSecondLevelModel}{$sl$, $data$}
		\State $\vartriangleright$ This trains a second-level model, where $sl$ is the model to be trained and $data$ is the data on which the first-level models are to be trained.
		
		\State
		\State $models$ = [ ]
		\State $flacc$ = [ ]
		\State $flcorr$ = [ ]
		\State \For{$model$ in $models$}
		\State $trained$ $\Leftarrow$ \Call{train}{$model$, $data$}
		\State $flacc$ append $trained.$\Call{score}{}
		\State $explanations$ = $trained.$\Call{getExplanations}{}
		\State $flcorr$ append \Call{compareExplanations}{$explanations$}
		\EndFor
		\State
		\State $A$ [ $accuracies$, $scores$] $\Leftarrow$ $flacc$, $flcorr$

		\State $second$ $\Leftarrow$ \Call{trainModel}{$sl$, $A$}

		\State \Return $second.$\Call{Score}{}
		\EndFunction
		\State
		\Function{compareExplanations}{$explanationList$}
		\State $\vartriangleright$ Compares results of multiple explanatory systems. $explanationList$ is a list of CSV filenames, created by \Call{getExplanations}{}
		\State
		
		\State $B$ $\Leftarrow$ read first file in $explanationList$
		\State $avgCorrelation$ = 0
		\For {$feature$ in $B$[$features$]}
		\State $avgCorrelation$ $\mathrel{+}=$ \Call{compareExplanationsSingleF}{$feature$, $explanationList$}
		\EndFor
		\State
		\Return $avgCorrelation$ / length of $B$[$features$]
		\EndFunction
		\State
		\Function{compareExplanationsSingleF}{$feature$, $explanationList$}
		\State $\vartriangleright$ Compares results of multiple explanatory systems. $explanationList$ is a list of CSV filenames, created by \Call{getExplanations}{} for a single feature{}
		\State
		\State$featureScore$ = [ ]
		\For {$file$ in $explanationList$}
		\State $B$ $\Leftarrow$ read $file$
		\For {$i$ in range length of $B$[$features$]}
		\State $featureScore$ append $B$[$scores$][$i$][index of $B$[$features$][$i$] that has $feature$]
		\EndFor
		\EndFor
		\State
		\State $data$ [$B$ [$explainer$]] $\Leftarrow$ $featureScore$
		\State
		\Return $data.$\Call{Corr}{}
		\EndFunction
		
	\end{algorithmic}
\end{algorithm}

\pagebreak
While the \Call{getExplanations}{} is not a contribution of this paper, we give a representative image of its output below, as we believe it will help in understanding the algorithm. We use the function provided by OmniXAI library's utilities as an example. 

\begin{figure}[H]
	\centering
	\includegraphics[width=0.8\textwidth]{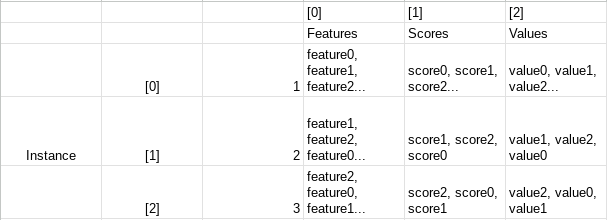}
	\caption{An example explanation file generated by \Call{getExplanations}{}. An example of how to read this file would be ``$feature0$ has an importance of $score0$ and value $value0$" For each instance, features are arranged by descending importance.}
	\label{fig:examplecorrplot2}
\end{figure}

\subsubsection{Does using maximum importance scores per instance instead of feature-specific importance scores provide more consistent correlation between explanatory systems?}
We hypothesize that if we use the maximum importance score, i.e. the magnitude of the score of the most important feature in each instance, the explanatory results will be more consistent than if we use the average correlation across all features. We test our hypothesis by following the preliminary procedure and the following step:

\begin{enumerate}
	\item For each dataset, compare the variance of the list of correlations for only the most important feature for each model versus the variance for the average correlation of feature importance scores across all features for these same models. 
\end{enumerate}

Repeat the above steps \emph{m} times to ensure consistency of results. Recall from the preliminary steps that we took \emph{m} = 5.

\section{Results}
Because we cannot include all generated graphs, we use this space to show the visuals and numbers we feel are the most important in supporting the conclusions of the paper. All data and figures can be found in Appendix~\ref{secA3}. Furthermore, since the results of all 10 of our trials were identical, we present the results of a single representative trial.

\begin{figure}[H]
	\centering
	\includegraphics[width=0.6\textwidth]{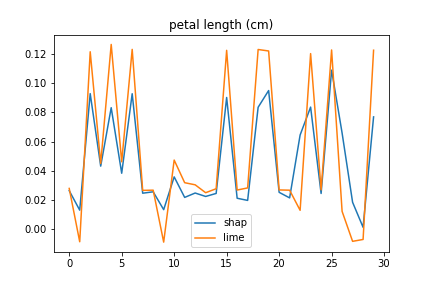}
	\caption{An example of a case where LIME and SHAP explainer scores are very similar. Here, we consider the feature petal length in the iris sk-learn dataset using Multionomial Naive Bayes. The correlation between the SHAP and LIME scores is in the range 0.9-1.0}
	\label{fig:examplecorrplot3}
\end{figure}

\begin{figure}[H]
	\centering
	\includegraphics[width=0.6\textwidth]{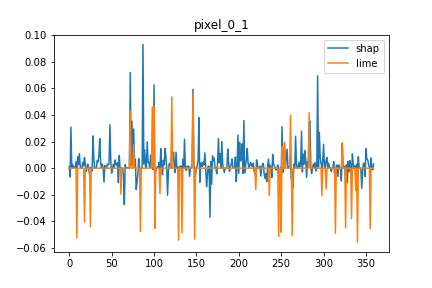}
	\caption{An example of a case where LIME and SHAP explainer scores differ by a wide margin. Here, we consider the feature pixel01 in the digits sk-learn dataset image classification task using Multionomial Naive-Bayes. The correlation between SHAP and LIME scores is less than 0.8.}
	\label{fig:examplecorrplot4}
\end{figure}

\begin{figure}[H]
	\centering
	\includegraphics[width=0.6\textwidth]{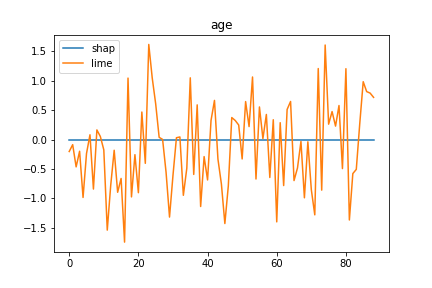}
	\caption{An example of a case where the correlation between LIME and SHAP explainer scores is undefined. While SHAP shows that the feature age is not important in predicting diabetes, LIME accords it some importance. From the diabetes regression task using Elastic Net.}
	\label{fig:examplecorrplotNan}
\end{figure}


In the correlation heatmaps below, each column is a feature (not labeled). 

\begin{figure}[H]
	\centering
	\includegraphics[width=0.6\textwidth]{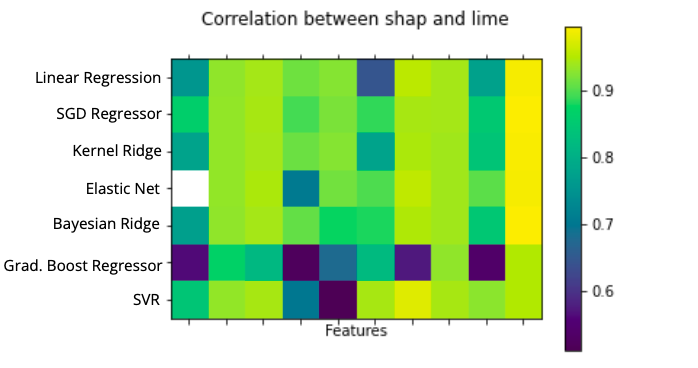}
	\caption{A correlation heat map for SHAP and LIME explanations for features in the sk-learn diabetes dataset, using different regression models.}
	\label{fig:examplecorrplot5}
\end{figure}
\begin{figure}[H]
	\centering
	\includegraphics[width=0.6\textwidth]{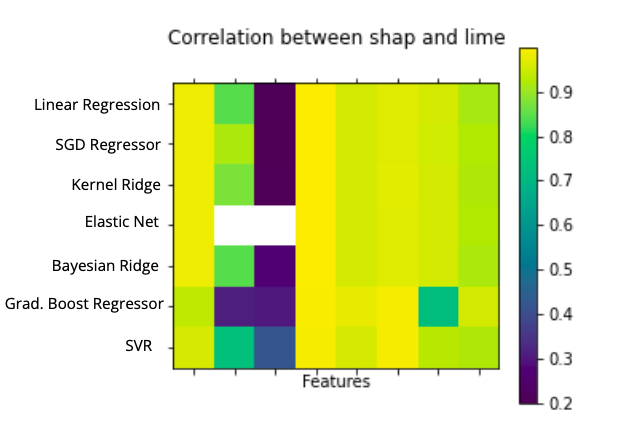}
	\caption{A correlation heat map for SHAP and LIME explanations for features in the UC Irvine Energy Efficiency dataset, using different regression models.}
	\label{fig:examplecorrplot6}
\end{figure}
\begin{figure}[H]
	\centering
	\includegraphics[width=0.2\textwidth]{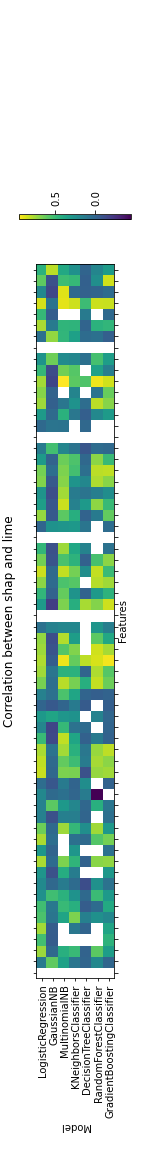}
	\caption{A correlation heat map for SHAP and LIME explanations for features in the sk-learn digits dataset, using different classification models.}
	\label{fig:examplecorrplot7}
\end{figure}
\begin{figure}[H]
	\centering
	\includegraphics[width=0.6\textwidth]{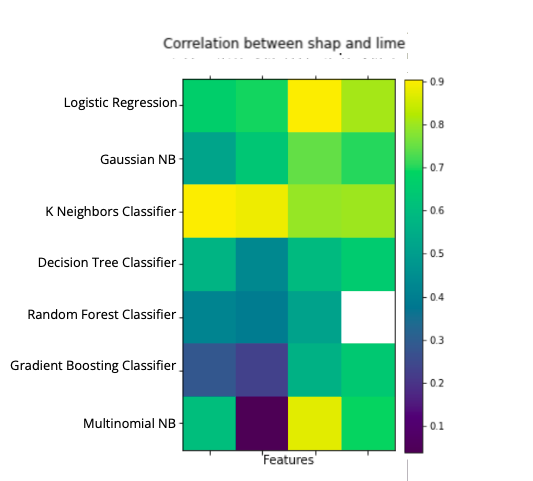}
	\caption{A correlation heat map for SHAP and LIME explanations for features in the sk-learn iris dataset, using different classification models.}
	\label{fig:examplecorrplot8}
\end{figure}

Below, we present the correlation and accuracy data used in analyzing the results of our study. These values are rounded to three decimal places.

Note: An \textbf{X} in Tables 2-5 indicates that the correlation was undefined. For the purpose of this paper, we take undefined correlations to be equivalent to zero correlation. Since the focus is on the relationship between model and correlation and not on features, the specific feature names are not listed in this table. Each row in Tables 2-5 represents a particular feature.

\begin{table}[h]
	\begin{minipage}{350pt}
		\caption{SHAP-LIME correlations per feature for all models on diabetes regression dataset}\label{corrtab}%
		\begin{tabular}{@{}lllllll@{}}
			\toprule
			Linear & SGD   & Kernel Ridge & Elastic Net & Bayesian Ridge & Gradient Boosting & SVR   \\
			\midrule
			0.757  & 0.862 & 0.779        & \textbf{X}  & 0.774          & 0.563             & 0.844 \\
			0.932  & 0.933 & 0.933        & 0.933       & 0.931          & 0.867             & 0.934 \\
			0.943  & 0.945 & 0.943        & 0.947       & 0.943          & 0.817             & 0.945 \\
			0.914  & 0.896 & 0.913        & 0.705       & 0.909          & 0.521             & 0.699 \\
			0.926  & 0.919 & 0.925        & 0.915       & 0.878          & 0.679             & 0.510 \\
			0.646  & 0.887 & 0.779        & 0.899       & 0.882          & 0.823             & 0.945 \\
			0.955  & 0.944 & 0.947        & 0.957       & 0.949          & 0.574             & 0.976 \\
			0.943  & 0.943 & 0.941        & 0.941       & 0.940          & 0.931             & 0.945 \\
			0.774  & 0.852 & 0.841        & 0.905       & 0.847          & 0.537             & 0.929 \\
			0.993  & 0.997 & 0.994        & 0.992       & 0.997          & 0.949             & 0.950 \\
			\botrule
		\end{tabular}
	\end{minipage}
\end{table}

\begin{table}[H]
	\begin{minipage}{350pt}
		\caption{SHAP-LIME correlations per feature for all models on energy-efficiency regression dataset}\label{corrtab2}%
		\begin{tabular}{@{}lllllll@{}}
			\toprule
			Linear & SGD   & Kernel Ridge & Elastic Net & Bayesian Ridge & Gradient Boosting & SVR   \\
			\midrule
			0.981  & 0.982 & 0.981        & 0.982       & 0.981          & 0.935             & 0.957 \\
			0.846  & 0.917 & 0.874        & \textbf{X}  & 0.845          & 0.309             & 0.728 \\
			0.210  & 0.199 & 0.210        & \textbf{X}  & 0.271          & 0.305             & 0.421 \\
			1.000  & 1.000 & 1.000        & 1.000       & 1.000          & 0.995             & 0.992 \\
			0.954  & 0.954 & 0.954        & 0.953       & 0.955          & 0.977             & 0.959 \\
			0.968  & 0.968 & 0.969        & 0.968       & 0.968          & 0.991             & 0.991 \\
			0.954  & 0.952 & 0.954        & 0.955       & 0.954          & 0.719             & 0.930 \\
			0.914  & 0.922 & 0.920        & 0.922       & 0.917          & 0.956             & 0.922 \\
			\botrule
		\end{tabular}
	\end{minipage}
\end{table}
\begin{longtable}{@{}lllllll@{}}
	\caption{SHAP-LIME correlations per feature for all models on digits classification dataset} 
	\label{corrtab3} \\
	\toprule
	Log.       & Gauss.NB   & KNeighbors & Dec.Tree   & Ran.Forest & GradBoost  & Multi.NB   \\
	\midrule
	\endfirsthead
	\caption{SHAP-LIME correlations per feature for all models on digits classification dataset (Continued) } 
	\\
	\toprule
	Log.       & Gauss.NB   & KNeighbors & Dec.Tree   & Ran.Forest & GradBoost  & Multi.NB   \\
	\midrule
	\endhead
	\textbf{X} & \textbf{X} & \textbf{X} & \textbf{X} & \textbf{X} & \textbf{X} & \textbf{X} \\
	0.115      & 0.610      & 0.351      & 0.092      & -0.000     & 0.191      & 0.005      \\
	0.741      & 0.014      & 0.432      & 0.501      & -0.012     & 0.599      & 0.326      \\
	0.525      & -0.053     & \textbf{X} & \textbf{X} & \textbf{X} & \textbf{X} & 0.454      \\
	0.757      & -0.045     & \textbf{X} & 0.100      & -0.006     & \textbf{X} & 0.364      \\
	0.553      & 0.085      & 0.363      & 0.677      & 0.155      & 0.248      & 0.193      \\
	0.522      & 0.107      & 0.496      & 0.431      & -0.046     & 0.032      & 0.401      \\
	0.250      & 0.521      & 0.450      & 0.257      & 0.022      & 0.506      & 0.303      \\
	0.207      & 0.013      & 0.125      & 0.029      & -0.128     & 0.252      & 0.069      \\
	0.279      & -0.113     & 0.414      & 0.131      & \textbf{X} & \textbf{X} & 0.287      \\
	0.777      & 0.030      & 0.676      & 0.467      & -0.020     & 0.701      & 0.722      \\
	0.288      & -0.006     & \textbf{X} & 0.281      & \textbf{X} & \textbf{X} & 0.070      \\
	0.775      & 0.046      & \textbf{X} & 0.062      & 0.029      & 0.579      & 0.662      \\
	0.642      & 0.024      & 0.749      & 0.480      & 0.236      & 0.756      & 0.291      \\
	0.156      & -0.112     & 0.147      & 0.144      & 0.000      & \textbf{X} & 0.052      \\
	0.158      & 0.598      & 0.292      & 0.188      & 0.012      & 0.434      & 0.086      \\
	0.192      & 0.026      & 0.059      & 0.003      & 0.124      & -0.464     & \textbf{X} \\
	0.046      & -0.081     & 0.118      & -0.007     & 0.232      & \textbf{X} & -0.015     \\
	0.837      & 0.008      & 0.676      & 0.669      & -0.113     & 0.731      & 0.751      \\
	0.820      & 0.029      & 0.618      & 0.510      & 0.125      & 0.765      & 0.709      \\
	0.776      & 0.079      & 0.759      & 0.436      & 0.100      & 0.738      & 0.815      \\
	0.496      & -0.157     & 0.820      & 0.372      & -0.025     & 0.119      & -0.016     \\
	0.185      & -0.165     & 0.488      & 0.070      & 0.032      & \textbf{X} & -0.004     \\
	0.294      & 0.366      & 0.284      & 0.259      & \textbf{X} & 0.164      & 0.003      \\
	-0.001     & -0.065     & -0.004     & 0.115      & -0.058     & \textbf{X} & -0.042     \\
	0.162      & 0.063      & 0.552      & 0.370      & -0.005     & -0.035     & -0.014     \\
	0.667      & 0.057      & 0.803      & 0.664      & 0.427      & 0.826      & 0.635      \\
	0.753      & 0.086      & 0.449      & 0.450      & -0.008     & 0.780      & 0.581      \\
	0.784      & -0.050     & 0.919      & 0.621      & 0.817      & 0.868      & 0.932      \\
	0.748      & -0.100     & 0.576      & 0.685      & \textbf{X} & 0.854      & 0.763      \\
	0.724      & -0.122     & 0.793      & 0.325      & \textbf{X} & 0.599      & 0.388      \\
	0.065      & 0.157      & 0.186      & 0.191      & \textbf{X} & 0.279      & 0.142      \\
	\textbf{X} & \textbf{X} & \textbf{X} & \textbf{X} & \textbf{X} & \textbf{X} & \textbf{X} \\
	0.301      & -0.206     & 0.676      & 0.428      & 0.771      & 0.693      & 0.850      \\
	0.468      & -0.011     & 0.622      & 0.266      & -0.010     & 0.304      & 0.455      \\
	0.708      & 0.030      & 0.552      & 0.579      & \textbf{X} & 0.800      & 0.756      \\
	0.840      & 0.029      & 0.828      & 0.660      & 0.230      & 0.825      & 0.507      \\
	0.652      & -0.094     & 0.549      & 0.454      & -0.023     & 0.544      & 0.709      \\
	0.474      & -0.114     & 0.746      & 0.379      & 0.168      & \textbf{X} & 0.072      \\
	\textbf{X} & \textbf{X} & \textbf{X} & \textbf{X} & \textbf{X} & \textbf{X} & \textbf{X} \\
	-0.004     & 0.262      & 0.275      & 0.295      & 0.086      & \textbf{X} & 0.036      \\
	0.729      & -0.062     & 0.596      & 0.424      & -0.012     & 0.113      & 0.617      \\
	0.324      & -0.079     & 0.841      & 0.158      & 0.304      & 0.713      & 0.493      \\
	0.267      & -0.123     & 0.759      & 0.128      & 0.085      & 0.146      & 0.231      \\
	0.687      & -0.065     & 0.696      & 0.274      & 0.193      & 0.779      & 0.706      \\
	0.661      & 0.049      & 0.675      & 0.534      & 0.055      & 0.808      & 0.820      \\
	0.312      & -0.013     & 0.637      & 0.558      & 0.016      & 0.707      & 0.497      \\
	0.126      & 0.527      & 0.184      & 0.074      & -0.087     & 0.036      & 0.021      \\
	\textbf{X} & \textbf{X} & \textbf{X} & \textbf{X} & \textbf{X} & \textbf{X} & \textbf{X} \\
	0.019      & 0.080      & 0.078      & \textbf{X} & 0.014      & \textbf{X} & \textbf{X} \\
	0.310      & 0.027      & 0.443      & 0.099      & -0.036     & 0.200      & 0.309      \\
	0.749      & -0.031     & 0.096      & 0.259      & 0.043      & 0.809      & 0.700      \\
	0.696      & -0.012     & \textbf{X} & 0.177      & \textbf{X} & 0.082      & 0.623      \\
	0.757      & -0.170     & 0.956      & 0.599      & 0.563      & 0.920      & 0.851      \\
	0.481      & -0.133     & 0.654      & 0.582      & \textbf{X} & 0.348      & 0.146      \\
	0.268      & 0.644      & 0.224      & 0.188      & 0.058      & 0.538      & 0.328      \\
	\textbf{X} & \textbf{X} & \textbf{X} & \textbf{X} & \textbf{X} & \textbf{X} & \textbf{X} \\
	0.059      & 0.726      & 0.472      & 0.344      & 0.040      & 0.058      & -0.033     \\
	0.776      & 0.116      & 0.466      & 0.458      & -0.009     & 0.448      & 0.472      \\
	0.486      & -0.037     & \textbf{X} & \textbf{X} & 0.114      & \textbf{X} & 0.019      \\
	0.862      & 0.078      & 0.897      & 0.847      & 0.507      & 0.843      & 0.834      \\
	0.201      & -0.164     & 0.888      & 0.019      & -0.025     & -0.007     & 0.039      \\
	0.604      & -0.112     & 0.516      & 0.487      & -0.028     & 0.236      & 0.856      \\
	0.447      & 0.809      & 0.391      & 0.249      & -0.006     & 0.182      & 0.317      \\
	\botrule
\end{longtable}
\begin{table}[H]
	\begin{minipage}{350pt}
		\caption{SHAP-LIME correlations per feature for all models on iris classification dataset}\label{corrtab4}%
		\begin{tabular}{@{}lllllll@{}}
			\toprule
			Log.  & Gauss.NB & KNeighbors & Dec.Tree & Ran.Forest & GradBoost & Multi.NB \\
			\midrule
			0.666 & 0.522    & 0.903      & 0.569    & 0.418      & 0.283     & 0.603    \\
			0.695 & 0.633    & 0.886      & 0.430    & 0.394      & 0.230     & 0.039    \\
			0.901 & 0.747    & 0.793      & 0.592    & 0.512      & 0.565     & 0.871    \\
			0.806 & 0.701    & 0.800      & 0.653    & \textbf{X} & 0.644     & 0.693    \\
			\botrule
		\end{tabular}
	\end{minipage}
\end{table}

In the tables and graphs below, \emph{Correlation Max} is referenced. This is the correlation between the maximum importance scores regardless of feature for each instance in a dataset. For the purpose of conciseness, we do not attach tables of instance-specific maximum importance scores, nor do we mention what features correspond to specific maximum importance scores. These tables and graphs were essential to our comparison of average explainer correlation with correlation of maximum importance scores and model accuracy. 

We now focus on each dataset.

\textbf{Diabetes Regression Analysis}
 
\begin{table}[H]
	\begin{minipage}{350pt}
		
		\caption{Diabetes regression: Low-performing models can still have high explainer correlations}\label{traintab1}%
		\begin{tabular}{@{}llll@{}}
			\toprule
			First Level Model & Avg. Correlation & Correlation Max & Accuracy \\
			\midrule
			Linear            & 0.878            & 0.335           & 0.332    \\
			SGD               & 0.918            & 0.832           & 0.323    \\
			Kernel Ridge      & 0.900            & 0.492           & -4.043   \\
			Elastic Net       & 0.819            & 0.851           & 0.357    \\
			Bayesian Ridge    & 0.905            & 0.844           & 0.332    \\
			Gradient Boosting & 0.726            & 0.743           & 0.203    \\
			SVR               & 0.868            & 0.805           & 0.128    \\
			\botrule
		\end{tabular}
	\end{minipage}
\end{table}
\begin{figure}[H]
	\centering
	\includegraphics[width = 1\textwidth]{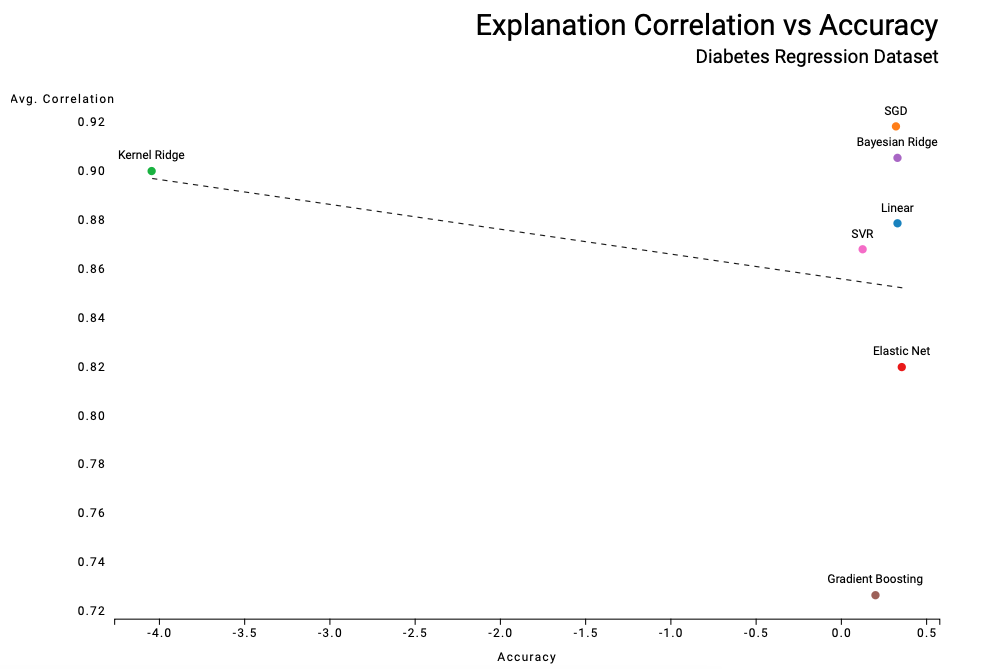}
	\caption{A scatterplot of SHAP-LIME correlation vs accuracy for different models on the diabetes regression task. The dashed line is the line of best fit.}
\end{figure}
\begin{figure}[H]
	\centering
	\includegraphics[width = 1\textwidth]{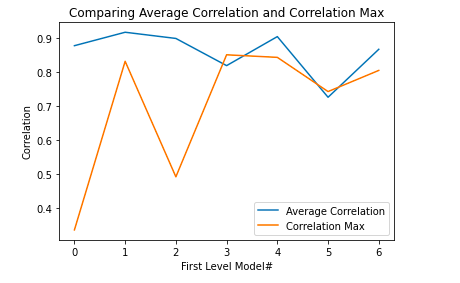}
	\caption{A line graph of average correlation and correlation between maximum importance scores}
\end{figure}

\begin{table}[H]
	\begin{minipage}{350pt}
		
		\caption{Training Second Level Models with Explanatory Correlations as Feature and Model Accuracy as Target for Diabetes Regression}\label{traintab2}%
		\begin{tabular}{@{}ll@{}}
			\toprule
			Second Level Model        & Accuracy \\
			\midrule
			Linear                    & -1.212   \\
			SGD                       & -0.981   \\
			KernelRidge               & -0.904   \\
			ElasticNet                & -1.182   \\
			BayesianRidge             & -1.207   \\
			GradientBoostingRegressor & -1.206   \\
			SVR                       & -1.152   \\
			\botrule
		\end{tabular}
	\end{minipage}
\end{table}

\pagebreak
\textbf{Energy Efficiency Regression Analysis}

\begin{table}[H]
	\begin{minipage}{350pt}
		\caption{Energy Efficiency regression: High-performing models may or may not have high explainer correlations.}\label{traintab3}%
		\begin{tabular}{@{}llll@{}}
			\toprule
			First Level Model & Avg. Correlation & Correlation Max & Accuracy \\
			\midrule
			Linear            & 0.853            & 0.666           & 0.909    \\
			SGD               & 0.862            & 1.000           & 0.905    \\
			Kernel Ridge      & 0.858            & 0.294           & -3.618   \\
			Elastic Net       & 0.723            & 0.989           & 0.799    \\
			Bayesian Ridge    & 0.861            & 0.449           & 0.908    \\
			Gradient Boosting & 0.773            & 0.854           & 0.997    \\
			SVR               & 0.863            & 0.732           & 0.917    \\
			\botrule
		\end{tabular}
	\end{minipage}
\end{table}
			
\begin{figure}[H]
	\centering
	\includegraphics[width = 1\textwidth]{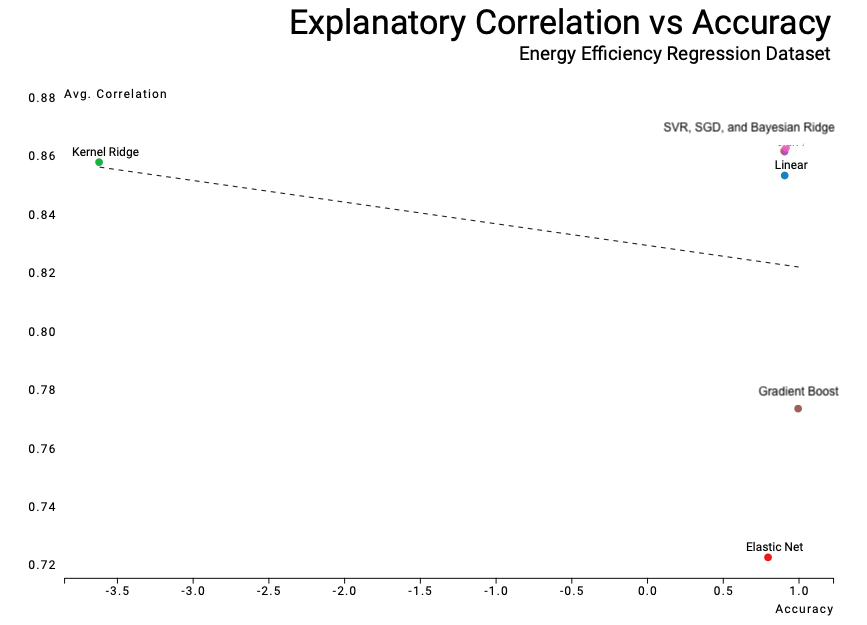}
	\caption{A scatterplot of SHAP-LIME correlation vs accuracy for different models on the energy efficiency regression task.The dashed line is the line of best fit.}
\end{figure}
\begin{figure}[H]
	\centering
	\includegraphics[width = 1\textwidth]{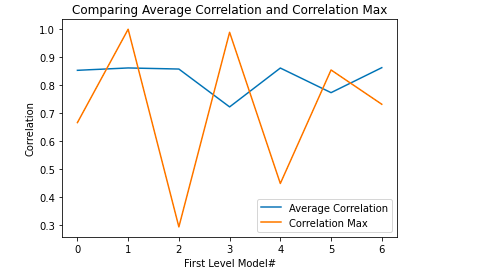}
	\caption{A line graph of average correlation and correlation between maximum importance scores}
\end{figure}

\begin{table}[H]
	\begin{minipage}{350pt}
		
		\caption{Training Second Level Models with Explanatory Correlations as Feature and Model Accuracy as Target for Energy Efficiency Regression}\label{traintab4}%
		\begin{tabular}{@{}ll@{}}
			\toprule
			Second Level Model        & Accuracy \\
			\midrule
			Linear                    & -1.003   \\
			SGD                       & -0.767   \\
			KernelRidge               & -0.362   \\
			ElasticNet                & -0.988   \\
			BayesianRidge             & -0.989   \\
			GradientBoostingRegressor & -0.992   \\
			SVR                       & -0.983   \\
			\botrule
		\end{tabular}
	\end{minipage}
\end{table}

\pagebreak
\textbf{Digits Classification Analysis}

\begin{table}[H]
	\begin{minipage}{350pt}
		
		\caption{Digits Classification: High-performing
			models may or may not have high explainer correlations}\label{traintab5}%
		\begin{tabular}{@{}llll@{}}
			\toprule
			First Level Model   & Avg. Correlation & Correlation Max & Accuracy \\
			\midrule
			Logistic Regression & 0.467            & 0.138           & 0.961    \\
			GaussianNB          & 0.0644           & 0.199           & 0.778    \\
			KNeighbors          & 0.515            & 0.210           & 0.969    \\
			DecisionTree        & 0.341            & 0.0167          & 0.842    \\
			RandomForest        & 0.100            & 0.0496          & 0.975    \\
			GradientBoosting    & 0.461            & 0.200           & 0.956    \\
			MultinomialNB       & 0.389            & 0.036           & 0.914    \\
			\botrule
		\end{tabular}
	\end{minipage}
\end{table}

\begin{figure}[H]
	\centering
	\includegraphics[width = 1\textwidth]{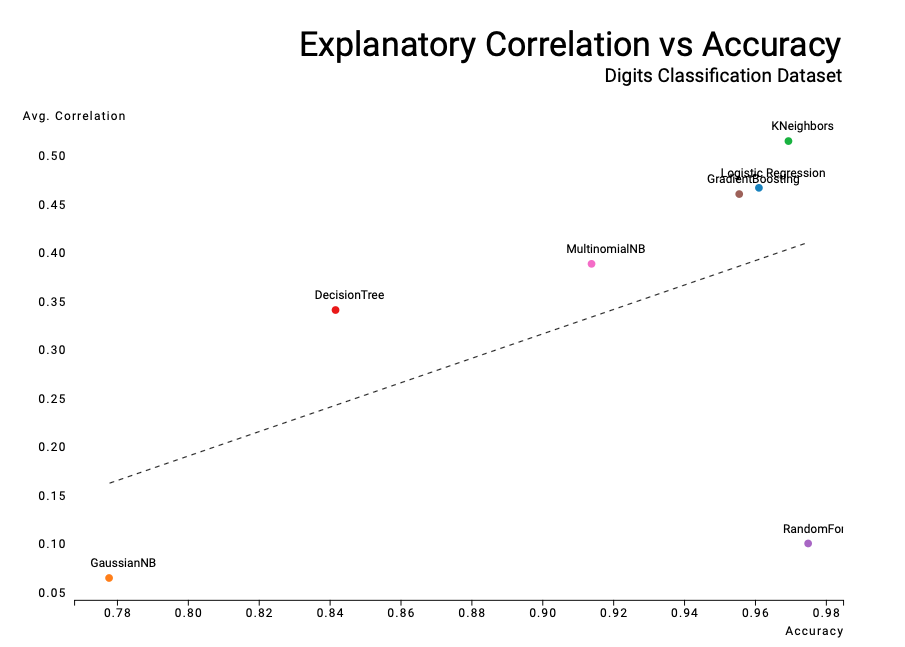}
	\caption{A scatterplot of SHAP-LIME correlation vs accuracy for different models on the digits classification task. The dashed line is the line of best fit.}
\end{figure}
\begin{figure}[H]
	\centering
	\includegraphics[width = 1\textwidth]{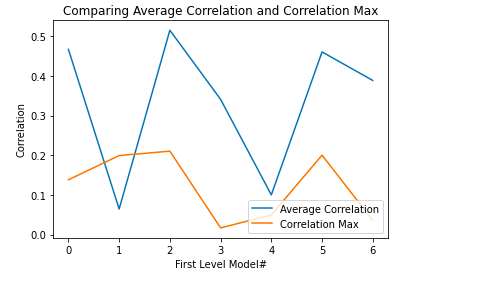}
	\caption{A line graph of average correlation and correlation between maximum importance scores}
\end{figure}

\begin{table}[H]
	\begin{minipage}{350pt}
		
		\caption{Training Second Level Models with Explanatory Correlations as Feature and Model Accuracy as Target for Digits Classification}\label{traintab6}%
		\begin{tabular}{@{}ll@{}}
			\toprule
			Second Level Model        & Accuracy  \\
			\midrule
			Linear                    & 0.675     \\
			SGD                       & -111.835  \\
			KernelRidge               & -1080.793 \\
			ElasticNet                & -2.016    \\
			BayesianRidge             & -1.326    \\
			GradientBoostingRegressor & -2.425    \\
			SVR                       & -5.522    \\
			\botrule
		\end{tabular}
	\end{minipage}
\end{table}

\textbf{Iris Classification Task}

\begin{table}[H]
	\begin{minipage}{350pt}
		
		\caption{Iris Classification: High-performing
			models may have very low explainer correlations}\label{traintab7}%
		\begin{tabular}{@{}llll@{}}
			\toprule
			First Level Model   & Avg. Correlation & Correlation Max & Accuracy \\
			\midrule
			Logistic Regression & 0.767            & 0.544           & 1.000    \\
			GaussianNB          & 0.651            & 0.613           & 0.967    \\
			KNeighbors          & 0.845            & 0.930           & 1.000    \\
			DecisionTree        & 0.561            & 0.384           & 1.000    \\
			RandomForest        & 0.442            & 0.444           & 1.000    \\
			GradientBoosting    & 0.431            & 0.155           & 1.000    \\
			MultinomialNB       & 0.552            & 0.279           & 0.567    \\
			\botrule
		\end{tabular}
	\end{minipage}
\end{table}

\begin{figure}[H]
	\centering
	\includegraphics[width = 1\textwidth]{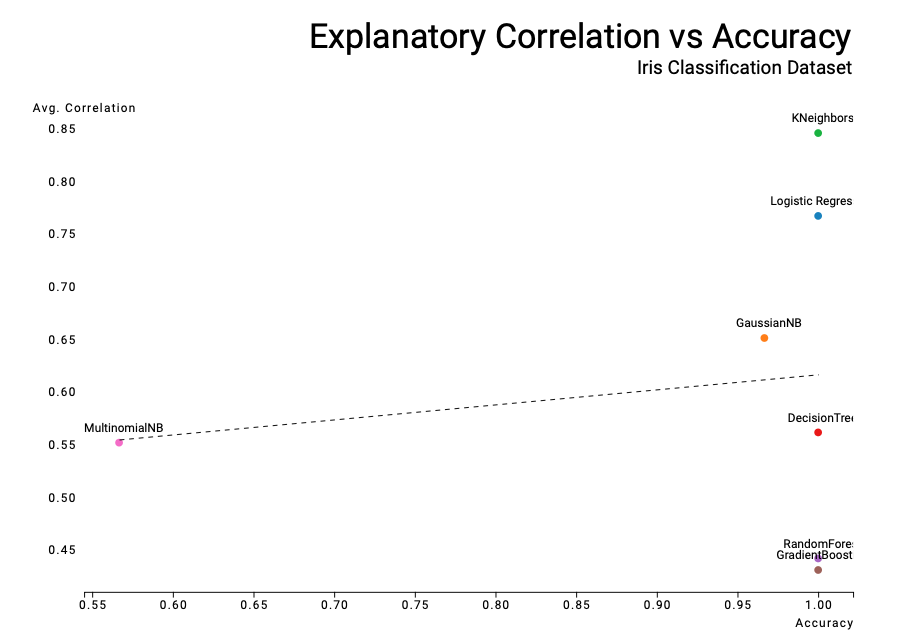}
	\caption{A scatterplot of SHAP-LIME correlation vs accuracy for different models on the iris classification task The dashed line is the line of best fit.}
\end{figure}
\begin{figure}[H]
	\centering
	\includegraphics[width = 1\textwidth]{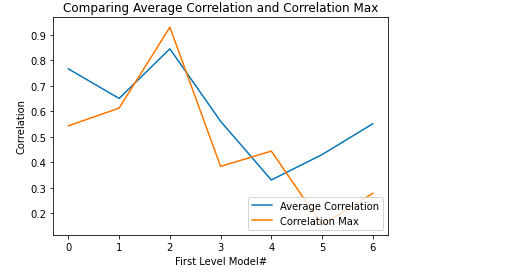}
	\caption{A line graph of average correlation and correlation between maximum importance scores}
\end{figure}

\begin{table}[H]
	\begin{minipage}{350pt}
		
		\caption{Training Second Level Models with Explanatory Correlations as Feature and Model Accuracy as Target for Iris Classification}\label{traintab8}%
		\begin{tabular}{@{}ll@{}}
			\toprule
			Second Level Model        & Accuracy \\
			\midrule
			Linear                    & -0.863   \\
			SGD                       & -0.092   \\
			KernelRidge               & -13.159  \\
			ElasticNet                & -0.939   \\
			BayesianRidge             & -0.919   \\
			GradientBoostingRegressor & -1.000   \\
			SVR                       & -0.852   \\
			\botrule
		\end{tabular}
	\end{minipage}
\end{table}

Below, we present a table of the variances of Avg Correlation and Correlation Max for all datasets. This data contradicts our hypothesis that Correlation Max will be less variable and therefore a more reliable indicator of SHAP-LIME correlations for all tasks.

\begin{table}[H]
	\begin{minipage}{350pt}
		\caption{Average Correlation Data Variance vs Correlation Max Variance - All datasets}\label{traintab9}%
		\begin{tabular}{@{}lll@{}}
			\toprule
			Dataset           & ${\sigma}_{AvgCorrelation}^2$ & ${\sigma}_{CorrelationMax}^2$ \\
			\midrule
			Diabetes          & 0.004                         & 0.042                         \\
			Energy Efficiency & 0.003                         & 0.071                         \\
			Digits            & 0.033                         & 0.007                         \\
			Classification    & 0.032                         & 0.063                         \\
			\botrule
		\end{tabular}
	\end{minipage}
\end{table}						

\section{Discussion}\label{sec13}

Both our hypotheses were not supported by the results of the study. Before we delve into the implications of the results, we first address some aspects of our data collection.

The undefined correlation values (indicated by \textbf{X}) occur when one explanatory system believes a feature is completely irrelevant to the model's predictions, while the other explanatory system disagrees. We leave further study of the implications of these undefined correlations to the readers. We do note however that some models, such as Elastic Net, are more likely to result in undefined correlations than others.

\textbf{Correlations predict model accuracy?} The results showed that correlation between SHAP and LIME importance scores cannot be used to deduce the accuracy of a given model, as the performance of all second-level models except for Linear Regression in the digits classification task was negative or otherwise below \emph{p} = 0.5. This is an important finding, which we use to postulate the following axiom that should be used for caution in future studies involving explanatory systems and machine learning models:

\begin{proposition}[Relationship between explanation correlation between two different explainers and model accuracy]\label{thm1}
	
	Let $I=(a, b)$ be an interval. Define $E_1$ and $E_2$ to be two explanatory monitors.   Let $acc$ be the accuracy achieved by some machine learning model over interval $I$.  Furthermore, let $\mathcal{I}_{E_1}$ and $\mathcal{I}_{E_2}$ be the importance scores of $E_1$ and $E_2$, respectively.
	
	Assume that $\mathcal{I}_{E_1}$ and $\mathcal{I}_{E_2}$ are given. We cannot infer $acc$ based on the correlation between $\mathcal{I}_{E_1}$ and $\mathcal{I}_{E_2}$.
\end{proposition}

\textbf{Use of maximum importance scores for each feature in a dataset presents a more consistent explanatory comparison method than using all importance scores?} The results showed that this is not true. Using only the maximum importance scores sometimes led to more variance in correlations between explanatory systems and therefore is not more reliable. This is an important finding, which we use to postulate the following axiom that should be used for caution in future studies involving explanatory systems and machine learning models:

\begin{proposition}[Correlation between maximum importance scores is not more consistent than correlation for all importance scores]\label{thm2}
	
	Consider $I_1$ and $I_2$, which are two importance scores defined over a set, $F$ of $n$ features: $F= \{f_1, f_2,...f_n\}$ for each instance.  Define $C_{i}$ as the correlation between the importance scores $I_1(f_i)$ and $I_2(f_i)$ across all instances for a particular feature $f_i$.  And let $C^*$ be the correlation between the maximum importance scores, $max(I_1)$ and $max(I_2)$, across all instances for each model.
	
	$\forall $i and $f_i \in F$, variance($C^*)$ is not necessarily less than $variance(C_i).$
\end{proposition}

\textbf{Additional Work} In addition to the XAISuite Framework and the result of our study, we would also like to briefly mention several related projects that we believe have relevance to a discussion about the XAISuite framework. Images of these tools in action can be found in Appendix \ref{secA5}. 

\begin{enumerate}
	\item{XAISuiteCLI: A comprehensive machine learning explainability command-line tool keeping with the XAISuite framework's emphasis on usability. This utility allows users to train and explain machine learning models using shell commands.}
	\item{XAISuiteGUI: A comprehensive graphical user interface that allows users without coding experience to train, explain, and compare machine learning models.}
	\item{XAISuiteBlock: A block-based site inspired by Scratch for machine learning model training and explanation. This is meant to offer machine learning utilities to those without coding experience. This is a great step forward in making machine learning explainability available to everyone regardless of age or coding experience.} We envision it as a potential educational tool.
\end{enumerate}

\textbf{Future Work} Our work opens up new areas of possible research. While we performed our analysis with 4 datasets and 7 models for each type of learning task, we encourage others to replicate our work with more datasets and models to confirm our results.
Furthermore, we understand that SHAP and LIME are inherently mathematical models, and we look forward to a mathematical basis for the results of our study. 
Finally, our goal is that the results of this paper, along with the framework we outline, will facilitate further efforts to resolve discrepancies between explanatory systems so that humans can gain a clearer understanding of how machine learning models work. This will lead, in turn, to a greater ability to fine tune these models to prevent error.

\backmatter

\bmhead{Supplementary information}

Additional files with all LIME and SHAP importance scores calculated by the authors for the purpose of this paper are included in Appendix~\ref{secA3}, except those that are presented in the Results section.

Additional graphs can also be found in Appendix~\ref{secA3}.

\section{Conclusions}
Explanatory systems allow users to look through the ``black box" of machine learning models. This is not only useful in understanding the internal mechanisms of machine learning models but also is essential in diagnosing model malfunctions. However, when multiple explanatory systems differ, users do not know which one to trust. 

This paper attempts to quantify and isolate scenarios leading to discrepancies between two widely used, open source machine learning explanatory systems, SHAP and LIME. By comparing importance scores for 14 models on a total of 4 varied tabular datasets, we gain important insights into the variability of importance scores and their implication on model accuracies. In the process, we construct the XAISuite framework, an important step towards making explanatory systems more accurate and understandable in the future.

Our paper finds that explanatory correlations cannot be used to predict model accuracy. Thus, if two explanatory systems differ, we cannot conclude that a model is necessarily less accurate, and vice versa. This has implications for user trust. Because users are often only supplied with model accuracy and explainer importance scores, this means that users do not have any information for determining which explanatory system is more accurate. 

Our paper also finds that the most important feature as determined by explanatory systems differs from explainer to explainer. More importantly, simply looking at the correlations between two explainers' top importance scores does not provide a more consistent way to compare explainers. Correlations vary from model to model and dataset to dataset, with no obvious pattern. 

Machine learning is a powerful tool. By arming users of machine learning with the information they need to make decisions, XAISuite increases trust. Furthermore, through its contribution of several interfaces catered to people regardless of age or coding experience, XAISuite empowers the use of machine learning among those that would be previously unable to do so. Finally, by setting the example for a comprehensive framework on machine learning explainability, XAISuite makes the entire process of machine learning more transparent and understandable. 

\section*{Declarations}

\begin{itemize}
	\item \textbf{Funding} Not applicable
	\item \textbf{Conflict of interest} Not applicable
	\item \textbf{Ethics approval} Not applicable
	\item {Consent to participate} Not applicable
	\item {Consent for publication} Not applicable
	\item \textbf{Availability of data and materials} All datasets used in this paper are on the public domain. All datasets except for the Energy Efficiency Dataset can be found on sk-learn's website. The Energy Efficiency Dataset can be found in UC Irvine's machine learning repository.\cite{energyefficiency}
	\item \textbf{Code availability} Any code used in this paper is the authors' own. See Appendix~\ref{secA2} for more details 
	\item \textbf{Authors' contributions} This paper was mainly written by Shreyan Mitra. Leilani Gilpin served in advisory capacity.
\end{itemize}

\noindent

\begin{appendices}
	
	\section{Elaborated Framework for Users}\label{secA1}
	Starting from the next page, we include a simplified version of our framework with helpful implementation tips.
	
	\includepdf[pages=1]{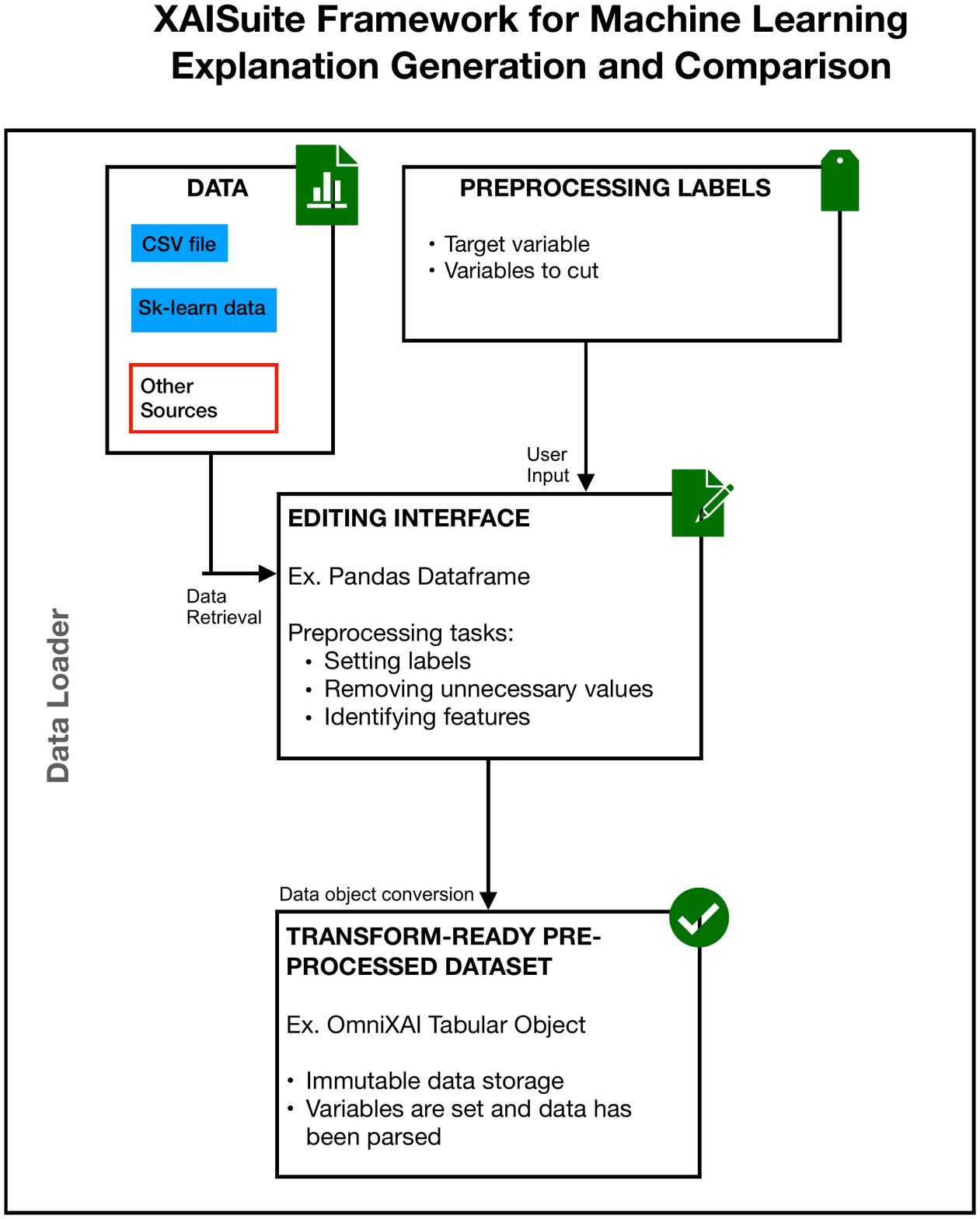}
	\includepdf[pages=2]{XAISuiteFlowchart.pdf}
	\includepdf[pages=3]{XAISuiteFlowchart.pdf}
	
	\section{Algorithm and Demo Code for Model Training and Explanation Generation (Online)}\label{secA2}
	
	An example of implementation for the XAISuite framework can be found in the XAISuite library public repository on Github: \hyperlink{https://github.com/11301858/XAISuite}{XAISuite}
	
	This link also contains additional utilities, like a command-line tool, a graphical user interface, and a block-coding site for beginners. Learn more at the \hyperlink{11301858.github.io/xaisuiteweb}{XAISuite website}.
	
	\section{Database, Supplementary Tables, and Graphs (Online)}\label{secA3}
	Go the link: \hyperlink{tinyurl.com/38myp2fc}{Full Experimental Results} This contains an entire database of collected importance scores, along with visuals that supplement this paper.
	
	\section{Additional Code}\label{secA4}
	\textbf{Setup Script}
	
	\begin{lstlisting}[language=Python, caption=Setup script for preliminary procedure]
from xaisuite import* # To install, pip install XAISuite
from sklearn import datasets*
#Energy_efficiency dataset will need to be uploaded to the working directory. The author used a version of the dataset found at https://www.kaggle.com/datasets/elikplim/eergy-efficiency-dataset

#Regression Tasks

models = ["LinearRegression", "SGDRegressor", "KernelRidge", "ElasticNet", "BayesianRidge", "GradientBoostingRegressor", "SVR"]

for j in range (len(models)):

    try:
        train_and_explainModel(models[j], load_data_CSV("energy_efficiency_data.csv", 'Heating_Load', 'Cooling_Load'), ["lime", "shap"], addendum = " energy_efficiency_data" )
    except:
        continue

data = ["load_diabetes()"]

for i in range (len(data)):
    for j in range (len(models)):
        
        try:
            train_and_explainModel(models[j], load_data_sklearn(eval(data[i]), 'target'), ["lime", "shap"], addendum = " " + data[i])
        except:
            continue

#Classification Task

models = ["LogisticRegression", "GaussianNB", "MultinomialNB", "KNeighborsClassifier", "DecisionTreeClassifier", "RandomForestClassifier", "GradientBoostingClassifier"]

data = ["load_digits()", "load_iris()"]
for i in range (len(data)):
    for j in range (len(models)):
        try:
            train_and_explainModel(models[j], load_data_sklearn(eval(data[i]), 'target'), ["lime", "shap"], addendum = " " + data[i])
        except ValueError:
            train_and_explainModel(models[j], load_data_sklearn(eval(data[i]), 'target'), ["lime", "shap"], scaleType = "MinMaxScaler", addendum = " " + data[i])
        except:
            continue
	\end{lstlisting}
	
	\section{Additional User Tools}\label{secA5}
	\textbf{XAISuiteCLI}
	\begin{figure}[H]
		\centering
		\includegraphics[width = 1\textwidth]{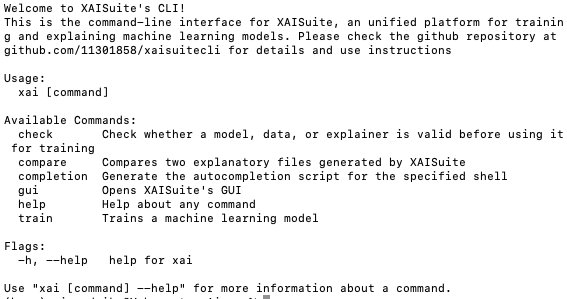}
		\caption{XAISuiteCLI's help menu output}
	\end{figure}
	\textbf{XAISuiteBlock}
	\begin{figure}[H]
		\centering
		\includegraphics[width = 1\textwidth]{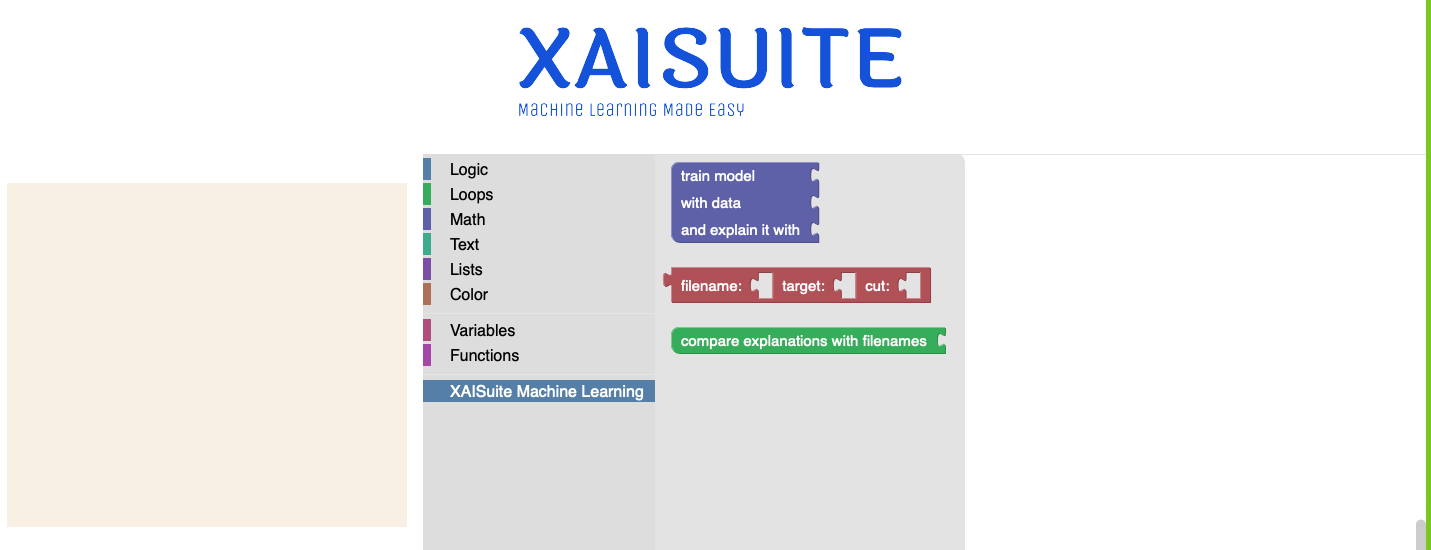}
		\caption{The options in XAISuiteBlock for training models}
	\end{figure}
	\textbf{XAISuiteGUI}
	\begin{figure}[H]
		\centering
		\includegraphics[width = 1\textwidth]{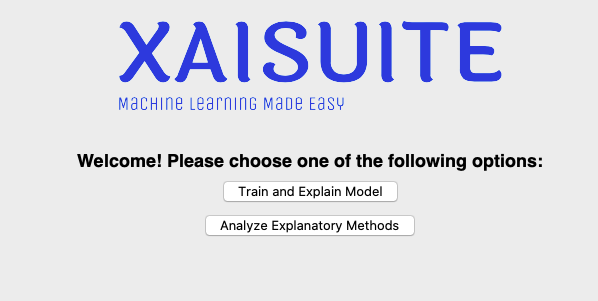}
		\caption{The home page of the XAISuiteGUI}
	\end{figure}
	\textbf{XAISuiteWeb}
	XAISuite's website provides access to the codebase, package, various user tools, and a coding playground to test XAISuite out. 
	\begin{figure}[H]
		\centering
		\includegraphics[width = 1\textwidth]{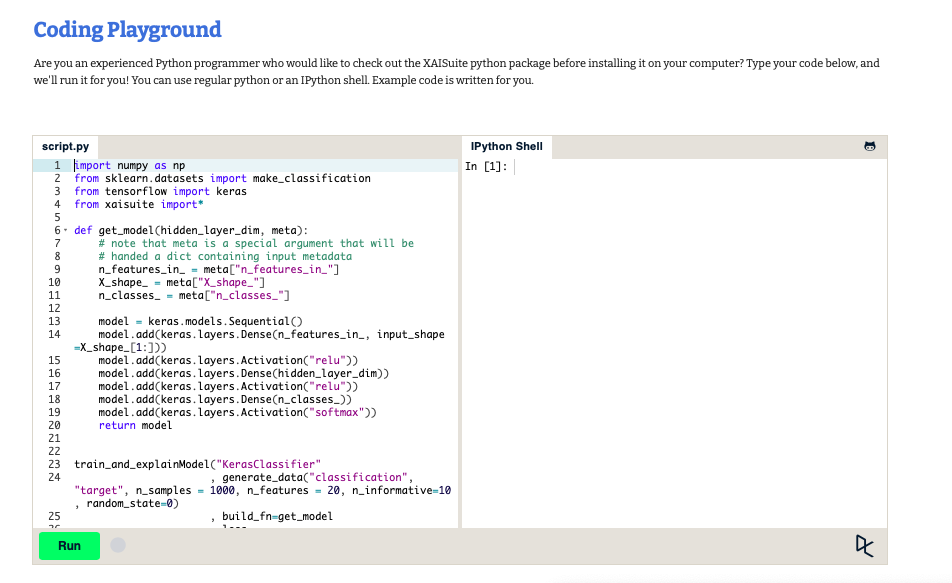}
		\caption{The XAISuite Coding Playground}
	\end{figure}
	    
	
	
	
\end{appendices}


\bibliography{sn-bibliography}


\end{document}